\documentclass[runningheads]{llncs}
\usepackage[T1]{fontenc}
\usepackage{graphicx,verbatim}
\usepackage{color}
\usepackage[hidelinks]{hyperref}
\usepackage{amsmath}
\usepackage{amssymb}
\usepackage{booktabs}
\usepackage{multirow}
\usepackage{pifont}
\usepackage{graphicx}
\usepackage{hhline}
\usepackage[table]{xcolor} 
\usepackage{enumitem}

\definecolor{rowgray}{gray}{0.90}
\definecolor{rowblue}{RGB}{225, 245, 250}
\newcommand{\cmark}{\textcolor{green!60!black}{\ding{51}}} 
\newcommand{\xmark}{\textcolor{red}{\ding{55}}}           
\definecolor{headergray}{gray}{0.80}

\begin{document}
\title{Hepato-LLaVA: An Expert MLLM with Sparse Topo-Pack Attention for Hepatocellular Pathology Analysis on Whole Slide Images}
%

\author{Yuxuan Yang$^{* \ 1}$, Zhonghao Yan$^{* \ \dagger \ 1}$, Yi Zhang$^{* \ 2}$, Bo Yun$^{1}$, \\ Muxi Diao$^{1}$, Guowei Zhao$^{2}$, Kongming Liang$^{\ddagger \ 1}$, Wenbin Li$^{\ddagger \ 2}$, Zhanyu Ma$^{1}$}  
\authorrunning{Yang et al.}
\institute{
    $^1$ School of Artificial Intelligence, Beijing University of Posts and Telecommunications, Beijing 100876, China \\
    $^2$ Department of Pathology, National Cancer Center/National Clinical Research Center for Cancer/Cancer Hospital, Chinese Academy of Medical Sciences and Peking Union Medical College, Beijing 100021, China \\
    \email{zhonghao.yan@bupt.edu.cn, zhangyi@cicams.ac.cn} \\
    $^*$Equal Contribution \quad $^\dagger$Project Lead \quad $^\ddagger$Corresponding Author
}
  
\maketitle              
\begin{abstract}
Hepatocellular Carcinoma diagnosis relies heavily on the interpretation of gigapixel Whole Slide Images. However, current computational approaches are constrained by fixed-resolution processing mechanisms and inefficient feature aggregation, which inevitably lead to either severe information loss or high feature redundancy. To address these challenges, we propose Hepato-LLaVA, a specialized Multi-modal Large Language Model designed for fine-grained hepatocellular pathology analysis. We introduce a novel Sparse Topo-Pack Attention mechanism that explicitly models 2D tissue topology. This mechanism effectively aggregates local diagnostic evidence into semantic summary tokens while preserving global context. Furthermore, to overcome the lack of multi-scale data, we present HepatoPathoVQA, a clinically grounded dataset comprising 33K hierarchically structured question-answer pairs validated by expert pathologists.
Our experiments demonstrate that Hepato-LLaVA achieves state-of-the-art performance on HCC diagnosis and captioning tasks, significantly outperforming existing methods. Our code and implementation details are available at \textcolor{blue}{\href{https://pris-cv.github.io/Hepto-LLaVA/}{\texttt{https://pris-cv.github.io/Hepto-LLaVA/}}}.


\end{abstract}

\section{Introduction}
Hepatocellular Carcinoma (HCC), a leading cause of global cancer mortality, relies on histopathological Whole Slide Images (WSIs) examination as the gold standard. However, HCC’s high heterogeneity and complex microenvironment render gigapixel WSIs interpretation labor-intensive. Manual analysis is prone to inter-observer variability, especially for subtle early-stage lesions.
Current Multiple Instance Learning (MIL) approaches~\cite{fourkioti2023camil}~\cite{song2024morphological}~\cite{wu2025learning}~\cite{zhuang2025libra} are often limited to classification, lacking clinical diagnostic capacity.
This has catalyzed WSI-based Multi-modal Large Language Models (MLLMs) \cite{seyfioglu2024quilt}~\cite{zhang2025patho}~\cite{chen2025slidechat}~\cite{liang2025wsi}~\cite{yan2025medreasoner}, which align visual pathology features with language to enable VQA.

A key design choice in these MLLMs is how to represent gigapixel WSIs for MLLMs. Existing methods typically fall into two categories: thumbnail-based approaches~\cite{seyfioglu2024quilt}~\cite{zhang2025patho} that resize the gigapixel image into a megapixel image, and slide-encoder-based approaches~\cite{chen2025slidechat}~\cite{liang2025wsi} that aggregate thousands of patches into global tokens.
However, thumbnail-based approaches inevitably lose patch-level details, while both methods ingest only WSIs inputs, consequently less equipped with patch-level capabilities essential for diagnosis. We investigate two primary research questions regarding the slide encoder, which acts as a bottleneck:

\begin{itemize}[leftmargin=4mm]
    \item \textbf{RQ1:} How can the slide encoder compress HCC WSIs into representations that preserve critical diagnostic details while minimizing redundancy?
    \item \textbf{RQ2:} How can the slide encoder accommodate variable-resolution inputs to generate features for multi-scale diagnosis in complex liver tissues?
\end{itemize}

To address \textbf{RQ1}, we design a \textbf{Sparse Topo-Pack Attention} mechanism that simulates the local aggregation and global juxtaposition of pathological diagnosis. By modeling 2D tissue topology, the slide encoder extracts topology-informed representations while significantly reducing spatial redundancy. A light-weight connector then condenses these features into fixed-length query tokens, effectively bridging the modality gap without losing critical diagnostic details.
To address \textbf{RQ2}, we leverage hierarchical features to develop \textbf{HepatoPathoVQA}, a comprehensive multi-scale VQA dataset specifically for HCC. This dataset contains over 33K expert-validated QA pairs covering the full clinical workflow across three distinct scales: WSIs, Region of Interest (ROI, 5$\times$ mag.), and Patch (10$\times$ and 20$\times$ mag.). Finally, through Low-Rank Adaptation (LoRA) fine-tuning on this unified framework, we present \textbf{Hepato-LLaVA}, a specialized MLLM optimized for fine-grained hepatocellular pathology analysis.

In summary, our three main contributions are:
\begin{enumerate}
\item We construct \textbf{HepatoPathoVQA}, the first multi-scale WSI dataset for HCC, comprising over 33K QA pairs across three scales, which improves multi-scale modeling and bridges data with real-world clinical practice.
\item We introduce a \textbf{Sparse Topo-Pack Attention} mechanism that models 2D tissue topology, extracting global-context-aware representations for WSIs while mitigating information redundancy.
\item We present \textbf{Hepato-LLaVA}, a specialized MLLM optimized via a three-stage pipeline, achieving an improvement of 20\% in average diagnostic accuracy in HCC over existing open-source pathology MLLMs.
\end{enumerate}

\section{Methods}

\subsection{Preliminary}

\noindent\textbf{Patch Encoding.}
Let $\mathbf{I}_{raw} \in \mathbb{R}^{H_{raw} \times W_{raw} \times 3}$ denote raw WSIs. We tessellate the tissue region into non-overlapping patches of size $P \times P$. This forms a grid layout of dimensions $H \times W$, where $H = \lfloor H_{raw}/P \rfloor$ and $W = \lfloor W_{raw}/P \rfloor$. Let $x_{i,j}$ represent the patch at the spatial coordinate $(i,j)$ for $1 \le i \le H, 1 \le j \le W$.
A frozen feature encoder $f_{enc}$ then maps each patch into a $D$-dimensional embedding space, yielding a feature grid:
\begin{equation}
    \mathbf{H}_{grid} = \left\{ \mathbf{h}_{i,j} \in \mathbb{R}^D \mid \mathbf{h}_{i,j} = f_{enc}(x_{i,j}) \right\}_{1 \le i \le H, \, 1 \le j \le W}
\end{equation}

\noindent\textbf{Slide Encoding.} 
Since the raw feature $\mathbf{H}_{grid}$ contains thousands of patches, directly utilizing it for slide-level tasks introduces excessive dimensionality and redundant information. To address this, a slide encoder $\mathcal{T}(\cdot)$ is employed to extract semantic features from the flattened patch sequence $\mathbf{S}_{patch} = \text{Flatten}(\mathbf{H}_{grid})$:
\begin{equation}
    \mathbf{H}_{slide} = \mathcal{T}(\mathbf{S}_{patch})
\end{equation}
In conventional MIL, $\mathcal{T}$ acts as a global aggregation function, where the output $\mathbf{H}_{slide} \in \mathbb{R}^{1 \times D}$ is a single compact token. In contrast, for Transformer-based approaches, $\mathcal{T}$ typically functions as a self-attention network where the output $\mathbf{H}_{slide} \in \mathbb{R}^{N \times D}$, often requiring a subsequent pooling to obtain a global token.

\subsection{HepatoPathoVQA Dataset}
\label{sec:HepatoPathoVQA}

\begin{figure}[t]
  \centering
  \includegraphics[width=\linewidth]{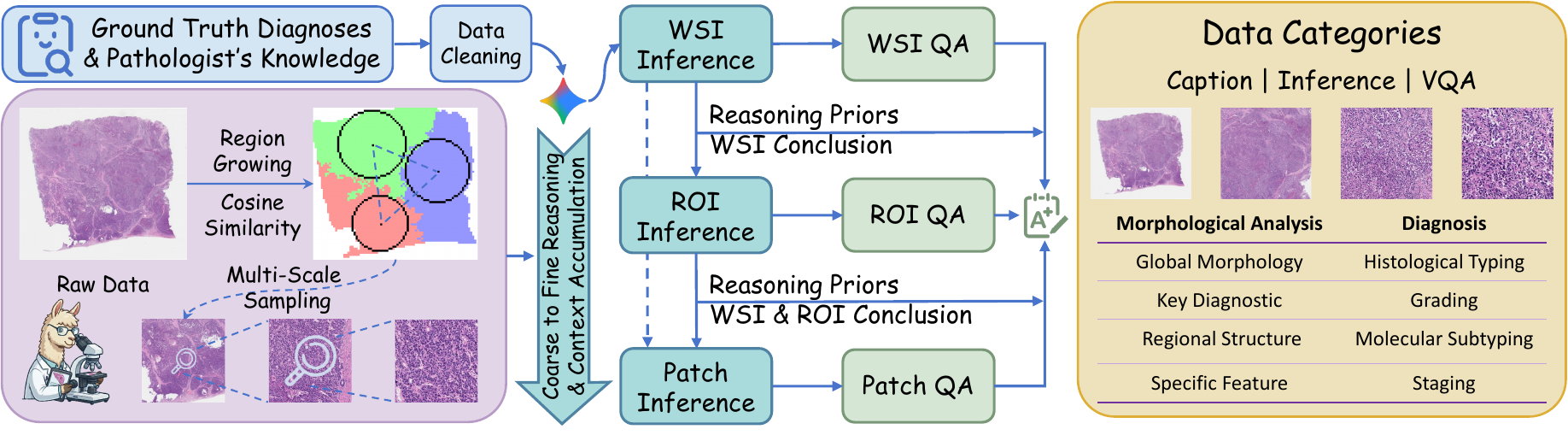}  
    \caption{Overview of the \textbf{HepatoPathoVQA} construction pipeline: (1) Extracts ROIs and Patches from WSIs using MST-based clustering and triangular seed-point selection. (2) Employs Gemini-3-flash for hierarchical inference by integrating macroscopic descriptions as context for subsequent microscopic analysis. (3) Generates multi-scale QA pairs and captions for instruction tuning and alignment.}
  \label{fig:data}
\end{figure}

\noindent\textbf{Dataset Construction.} We collected 200 WSIs containing HCC, with all patient identifiers removed to ensure privacy. From these WSIs, we constructed \textbf{HepatoPathoVQA}, a multi-scale dataset featuring 33K QA pairs centered on morphological analysis and diagnosis. As shown in Fig. ~\ref{fig:data}, following the standard diagnostic workflow of pathologists, we developed a data generation pipeline using Gemini-3-flash~\cite{google_gemini_2026}. This pipeline simulates the clinical reasoning process, transitioning from macroscopic observation to microscopic details. Each step in the pipeline is designed as an atomic operation.
The data construction pipeline consists of three stages. 
\textbf{(1) Hierarchical Multi-Scale Sampling.} WSIs serve as the raw input. We identify ROIs using a triangular seed-point algorithm and a Minimum Spanning Tree (MST). The MST aggregates adjacent patches based on cosine similarity to form three candidate regions. These ROIs represent a 5$\times$ magnification based on medical priors of cell adjacency. We manually remove uninformative regions. Finally, 10$\times$ and 20$\times$ local patches are randomly sampled around the ROI centers.
\textbf{(2) Context-Injected Reasoning Generation.} Since Gemini cannot process gigapixel images, all regions are resized to 2048$\times$2048 pixels. We design scale-specific prompts based on clinical focus and diagnostic status. Gemini generates descriptions in a hierarchical manner. Descriptions from higher scales serve as context for the next scale to ensure logical consistency from macroscopic to microscopic levels.
\noindent\textbf{(3) Scale-aware Data Synthesis.} Leveraging the generated hierarchical descriptions, we generate diverse multi-scale QA pairs for instruction tuning. These pairs cover all three resolutions, enabling the model to perform both granular morphological analysis and holistic clinical diagnosis. Furthermore, these descriptions are utilized to produce high-quality image-text captions, which serve as the foundation for connector pre-training to achieve robust vision-language alignment.

\noindent\textbf{Dataset Statistics. } \textbf{HepatoPathoVQA} comprises 33,332 QA pairs. At our three proposed spatial scales, we follow the WSI-LLaVA~\cite{liang2025wsi} setup to prompt the model to generate reasoning chains along two axes—morphology and diagnosis. Each dimension is further subdivided into four subcategories, thereby supporting hierarchical diagnosis from overall patterns to cellular details. Beyond the VQA pairs, we provide \textbf{HepatoPathoCaption} (3,288 pairs) for alignment pre-training and \textbf{HepatoPathoBench} (3,056 pairs) split from HepatoPathoVQA for evaluation. The entire dataset construction process was validated by three expert pathologists under a blind evaluation protocol, yielding high inter-rater consistency (Pearson $r = 0.96$) and a low rejection rate ($< 2\%$).

\subsection{Sparse Topo-Pack Attention}

\noindent\textbf{Topological Mismatch in Existing Modeling.}
Existing pathology-specific MLLM predominantly rely on Longnet to extract features, treating the WSIs as a flattened 1D sequence. This assumption neglects the intrinsic 2D topological properties of pathological tissues. Biologically, semantic dependencies vary significantly across scales: local regions share a common semantic meaning(e.g., tumor margins crossing multiple patches), whereas distant regions are distinct with lower semantic coupling. To address this, we propose a hierarchical sparse attention mechanism that restores topological priors (Fig.~\ref{fig:hepto}, upper panel).

\noindent\textbf{Hierarchical Sequence Construction.}
Unlike standard flattening, we construct a structured input sequence that explicitly encodes the grid hierarchy. 
We organize this grid into $M$ Summary Packs in row-major order. Each pack $\mathbf{P}_m$ represents a local $k \times k$ window.
We generate the pack summary token $\mathbf{s}_m$ by resizing the local image region of pack $m$ and passing it through the patch encoder $f_{enc}$ for semantic anchors. Similarly, the global token $\mathbf{g}_{global}$ is derived by resizing entire WSIs and encoding it.
The total sequence $\mathbf{S}_{in}$ is formed as:
\begin{equation}
    \mathbf{S}_{in} = [ \ \mathbf{g}_{global}, \ \underbrace{\mathbf{h}_{1,1}^1, \dots, \mathbf{h}_{k,k}^1, \mathbf{s}_1}_{\text{Pack } \mathbf{P}_1}, \ \dots, \ \underbrace{\mathbf{h}_{1,1}^M, \dots, \mathbf{h}_{k,k}^M, \mathbf{s}_M}_{\text{Pack } \mathbf{P}_M} \ ]
\end{equation}
where $\mathbf{h}_{i,j}^m$ represents the fine-grained patch tokens within the $m$-th pack. In our implementation, we set $k=3$, resulting in 10 tokens per pack. Through the hierarchical attention mechanism, the summary token functions as a dynamic query that aggregates diagnostic evidence from patch tokens $\mathbf{h}_{i,j}^m$.

\noindent\textbf{Hierarchical Sparse Mask.}
We define a hierarchical mask $\mathcal{M}$ to enforce the specific interaction rules. Let $\Omega_P$ and $\Omega_S$ denote the sets of all patch tokens and all summary tokens, respectively. Let $\mathcal{P}(t)$ denote the pack index of a token $t$.
For any query token $t_i$ and key token $t_j$ in $\mathbf{S}_{in}$, the attention mask $\mathcal{M}_{i,j}$ is:
\begin{equation}
    \mathcal{M}_{i,j} = \begin{cases} 
    0 & \text{if } t_j = \mathbf{g}_{global} \quad (\text{Global Sink}) \\
    0 & \text{if } t_i, t_j \in \Omega_P \land \mathcal{P}(t_i) = \mathcal{P}(t_j) \quad (\text{Intra-Pack Dense}) \\
    0 & \text{if } t_i \in \Omega_S \land t_j \in \Omega_P \land \mathcal{P}(t_i) = \mathcal{P}(t_j) \quad (\text{Aggregation}) \\
    0 & \text{if } t_i, t_j \in \Omega_S \quad (\text{Summary-Level Interaction}) \\
    -\infty & \text{otherwise}
    \end{cases}
\end{equation}
This hierarchical architecture effectively models pathological topology by harmonizing dual-scale interactions. The global token provides a macro-level reference, enabling local windows to focus on aggregating dense features into summary tokens. Summary tokens then facilitate long-range modeling, preserving the structural integrity of tissue regions across entire slide. In our implementation ($k=3$), this sparsity reduces the attention overhead to $\approx \mathbf{1\%}$ of the dense counterpart.

\subsection{Hepato-LLaVA}

\begin{figure}[t]
  \centering
  \includegraphics[width=\linewidth]{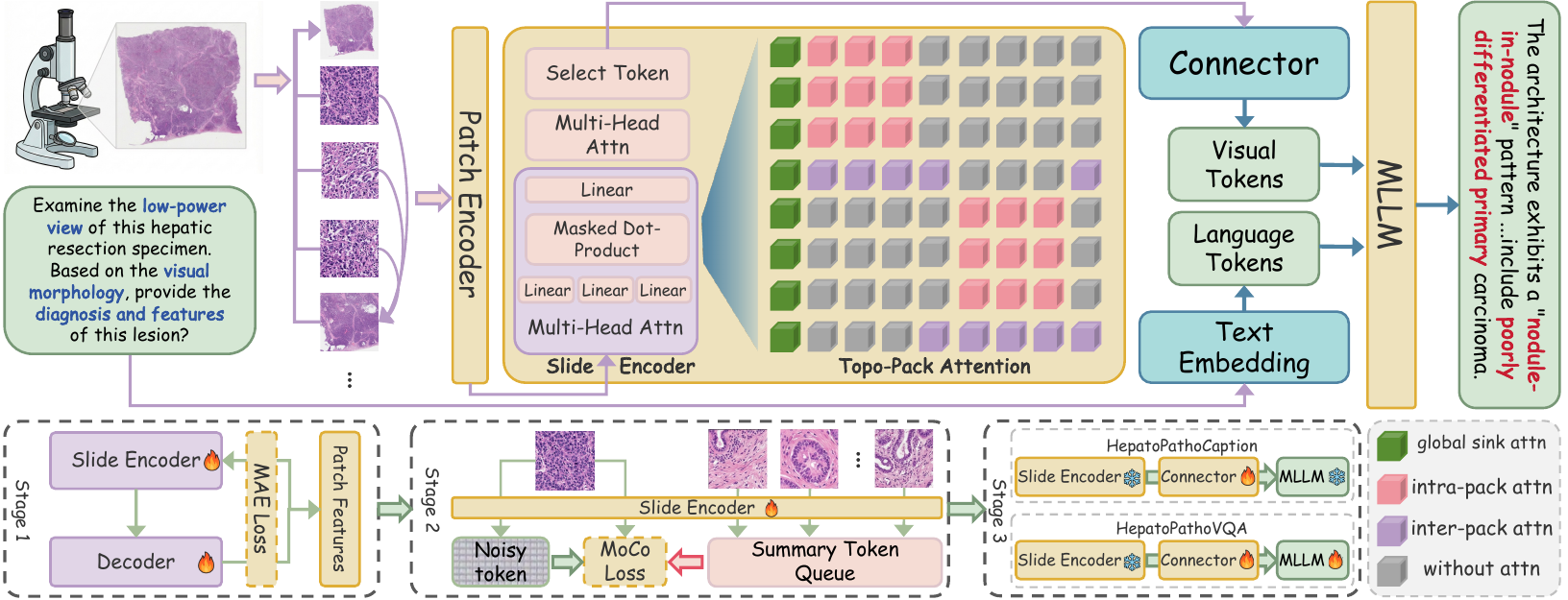}  
  \caption{Overview of the \textbf{Hepato-LLaVA} framework: (Upper) Incorporates Sparse Topo-Pack Attention into the model architecture. (Lower) Implements a three-stage training pipeline: MAE pre-training, MoCo pre-training, and instruction tuning (via LoRA). The sparse attention mask defines three topological interactions: (1) Global Sink for macro-context broadcasting, (2) Intra-Pack for local dense interactions, and (3) Inter-Pack for summary-level connections across packs.}
  \label{fig:hepto}
\end{figure}

\noindent\textbf{MAE Pretrain.}
As shown in the lower panel of Fig.~\ref{fig:hepto}, we applied MAE pre-training on a composite dataset consisting of TCGA~\cite{weinstein2013cancer} and our internal data to optimize feature extraction. The dataset was augmented via spatial chunking and geometric flipping. The core innovation lies in a two-stage curriculum masking strategy: \textbf{Phase 1} implements patch-wise masking to reconstruct missing patches within intact packs, focusing on capturing tissue textures; \textbf{Phase 2} advances to pack-wise masking, where entire packs are removed to enforce long-range dependency modeling, thereby capturing high-level structural patterns.

\noindent\textbf{MoCo Pretrain.}
Following MAE pre-training, this stage utilizes the Momentum Contrast (MoCo)~\cite{he2020momentum} architecture. We utilize the un-augmented dataset to prevent the inclusion of augmented views as negative samples. This stage aims to align visual representations at the \textbf{Summary Token level}, focusing on tissue semantics rather than holistic descriptors. Addressing the high I/O costs of WSIs processing, we adapt the MoCo framework to operate on feature-level, generating positive pairs by injecting noise directly into token features. Coupled with a momentum-updated encoder and a queue containing cross-case negative samples, the model is optimized via InfoNCE loss to discriminate specific regions.

\noindent\textbf{Multi-scale Instruction Tuning.}
To effectively bridge the modality gap between the pre-trained, frozen slide encoder and the MLLM, we introduce a Q-Former~\cite{li2023blip} Connector. To handle variable-length pack summary token sequences, this lightweight module employs a set of learnable query vectors to interact with inputs.
The instruction tuning process is further divided into two sub-phases: 
(1) Visual-Language Alignment: We first train the connector on the \textbf{HepatoPath-Caption} dataset while keeping both the slide encoder and MLLM frozen. This stage aligns visual representations with linguistic semantics, enabling the connector to translate pathological features.
(2) Diagnostic Instruction Tuning: We fine-tune both the connector and the MLLM on the \textbf{HepatoPath-VQA} dataset. This stage optimizes the model for diagnostic inference, enabling it to interpret multi-scale visual evidence and generate clinically precise responses.

\section{Experiments}
\subsection{Experimental Settings}
\noindent\textbf{Datasets.}
The slide encoder is pre-trained on 10K WSIs aggregated from TCGA \cite{weinstein2013cancer}, HCMI~\cite{hcmi}, and internal dataset. For instruction tuning, we utilize HepatoPathoCaption (3K pairs) and HepatoPathoVQA (33K pairs). Evaluation is performed on HepatoPathoBench (3K pairs), strictly disjoint from the training.

\noindent\textbf{Models.}
We conduct a comparison across three types of 7B models.
For general medical models, we evaluate HuatuoGPT~\cite{zhang2023huatuogpt} and Lingshu~\cite{xu2025lingshu}.
For thumbnail-based pathology MLLMs, we include Quilt-LLaVA~\cite{seyfioglu2024quilt} and Patho-R1~\cite{zhang2025patho}(inputs resized to their native square resolutions).
For WSI-based pathology MLLMs with sparse attention, we compare against SlideChat~\cite{chen2025slidechat} and WSI-LLaVA~\cite{liang2025wsi}.

\noindent\textbf{Metrics.}
To evaluate the clinical accuracy of open-ended responses, we employ METEOR~\cite{banerjee2005meteor} and WSI-P~\cite{liang2025wsi}, which leverage LLM to evaluate predictions, providing a more reliable proxy for pathologist expertise. For single-choice questions, we adopt standard Accuracy based on a strict match. For multiple-choice questions, predictions that are correct but incomplete are assigned a score of 0.5.

\begin{table}[t]
\centering
\setlength{\aboverulesep}{0pt}
\setlength{\belowrulesep}{0pt}
\renewcommand{\arraystretch}{1.3}

\caption{Evaluation on HepatoPathoBench. Baselines include general medical MLLMs and pathology-specific MLLMs. Single/Multi refer to single/multiple choice tasks. WSI, ROI, and Patch denote the spatial scales. The Avg score is calculated across all tasks. Best results are \textbf{bold}; second-best are \underline{underlined}}

\resizebox{\textwidth}{!}{
\begin{tabular}{l c cc cc cc cc ccc c}
\toprule

& & \multicolumn{4}{c}{\textbf{Morphological Analysis}} & \multicolumn{4}{c}{\textbf{Diagnosis}} & \multicolumn{3}{c}{\textbf{Multi-scale}} & \multirow{3}{*}{\textbf{Avg}} \\
\cmidrule(lr){3-6} \cmidrule(lr){7-10} \cmidrule(lr){11-13}

& & \multicolumn{2}{c}{Open} & \multicolumn{2}{c}{Close} & \multicolumn{2}{c}{Open} & \multicolumn{2}{c}{Close} & & & \\
\cmidrule(lr){3-4} \cmidrule(lr){5-6} \cmidrule(lr){7-8} \cmidrule(lr){9-10}

\multirow{-3}{*}{\textbf{Model}} & \multirow{-3}{*}{\textbf{Input}} &
WSI-P & METEOR & Single & Multi &
WSI-P & METEOR & Single & Multi &
\multirow{-2}{*}{WSI} & \multirow{-2}{*}{ROI} & \multirow{-2}{*}{Patch} \\

\midrule

Lingshu & Thumbnail & 
0.53 & 0.17 & 0.38 & 0.44 &
\underline{0.73} & 0.18 & 0.39 & 0.38 &
0.52 & 0.52 & 0.49 & 0.50 \\

Huatuo-GPT & Thumbnail & 
\underline{0.74} & \underline{0.24} & 0.81 & 0.45 &
0.70 & \underline{0.23} & 0.59 & 0.32 &
0.60 & 0.65 & 0.65 & 0.65 \\

Quilt-LLaVA & Thumbnail & 
0.64 & 0.22 & 0.47 & 0.32 &
0.56 & 0.15 & 0.57 & 0.37 &
0.57 & 0.60 & 0.55 & 0.57 \\

Patho-R1 & Thumbnail & 
0.66 & 0.19 & \underline{0.87} & \underline{0.50} &
0.20 & 0.05 & 0.59 & \underline{0.45} &
0.55 & 0.55 & 0.54 & 0.55 \\

SlideChat & WSI & 
0.70 & 0.17 & \underline{0.87} & 0.47 &
0.72 & 0.14 & 0.63 & 0.39 & 
\underline{0.66} & \underline{0.68} & \underline{0.66} & \underline{0.66} \\

WSI-LLaVA & WSI & 
0.69 & 0.20 & 0.84 & 0.46 &
0.67 & 0.16 & \underline{0.65} & 0.36 &
0.65 & 0.67 & 0.64 & 0.65 \\

\rowcolor{rowblue} \textbf{Hepato-LLaVA} & WSI & 
\textbf{0.79} & \textbf{0.33} & \textbf{0.97} & \textbf{0.88} &
\textbf{0.75} & \textbf{0.33} & \textbf{0.87} & \textbf{0.68} &
\textbf{0.82} & \textbf{0.83} & \textbf{0.83} & \textbf{0.83} \\

\bottomrule
\end{tabular}
}
\label{tab:performance}
\end{table}

\subsection{Main Results}
\begin{figure}[t]
  \centering
  \includegraphics[width=\linewidth]{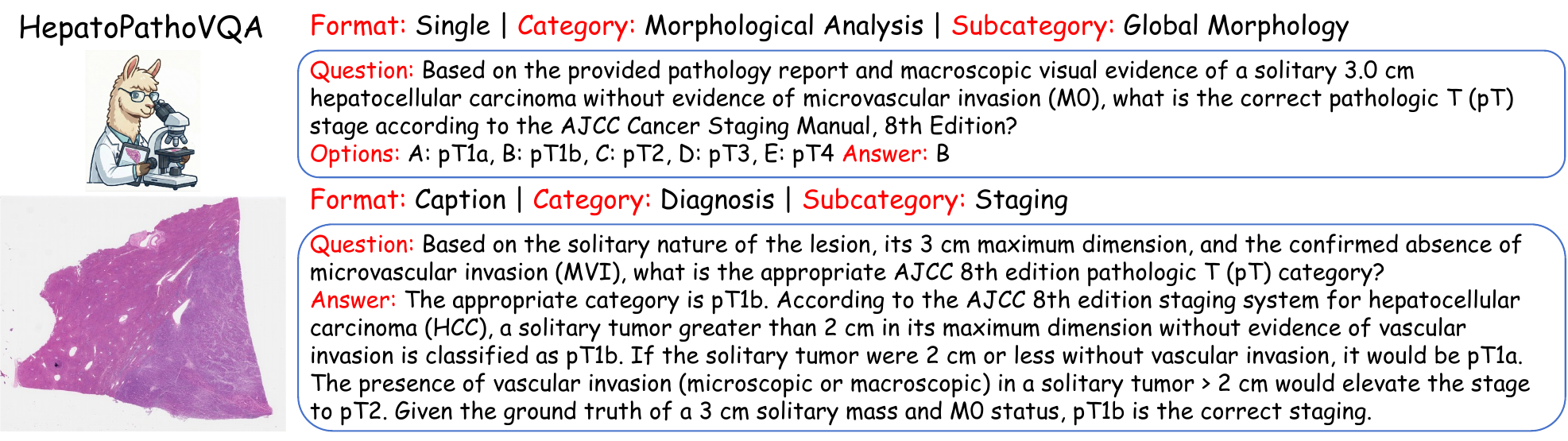}  
    \caption{A representative diagnostic case from HepatoPathoVQA, illustrating Hepato-LLaVA’s capability to provide precise by interpreting critical morphological evidence.}
  \label{fig:case}
\end{figure}

Hepato-LLaVA is initialized from WSI-LLaVA, which is fine-tuned using LoRA (r=128), while the connector is trained from scratch.
We evaluated Hepato-LLaVA on HepatoPathoBench against six baselines. As shown in Table~\ref{tab:performance}, while Thumbnail-based models yielded suboptimal results (Avg 0.50–0.57) due to severe information loss, Hepato-LLaVA achieved state-of-the-art performance (Avg \textbf{0.83}). This establishes a substantial lead of 0.17 over the best WSI-based runner-up SlideChat (0.66). In open-ended tasks, Hepato-LLaVA achieved a WSI-P of 0.79 in morphology and 0.75 in diagnosis, surpassing the SlideChat (0.70, 0.72). In close-ended tasks, our model attained 0.97 accuracy in morphological single-choice and 0.88 in multi-choice, significantly outperforming Patho-R1 (0.87, 0.50), a strong baseline enhanced by reinforcement learning. Regarding multi-scale consistency, Hepato-LLaVA demonstrated robust performance across WSI (0.82), ROI (0.83), and Patch (0.83) levels. This represents an improvement over our backbone model WSI-LLaVA (0.65, 0.67, 0.64), validating that the proposed sparse attention and connector effectively overcomes the scale variance limitations inherent in the base architecture.
Qualitative evidence in Fig.~\ref{fig:case} confirms its ability to interpret critical morphological evidence for precise diagnosis.

\subsection{Ablation Study}

\begin{table}[t]
    \centering
    \begin{minipage}[t]{0.48\textwidth}
        \rule{0pt}{0pt}
        \centering
        \centering

\setlength{\tabcolsep}{6pt}
\renewcommand{\arraystretch}{1.2}

\caption{Ablation of connector types and two-stage training strategy. For two-stage setting, we pre-train the connector on HepatoPathCaption with MLLM frozen.}

\resizebox{\textwidth}{!}{
\begin{tabular}{l c c c c}
\toprule
\textbf{Connector} & \textbf{Two-Stage} & \textbf{WSI} & \textbf{ROI} & \textbf{Patch} \\
\midrule

wsi-llava & \xmark & 81.61 & \underline{88.05} & \underline{86.83} \\
mlp & \xmark & 74.43 & 82.83 & 79.97 \\
mlp & \cmark & \underline{83.33} & 87.24 & 83.66 \\
qformer & \xmark & \textbf{88.51} & 87.82 & 85.34 \\
qformer & \cmark & \textbf{88.51} & \textbf{88.75} & \textbf{88.10} \\

\bottomrule
\end{tabular}
}
\label{tab:ablation_connector}
    \end{minipage}
    \hfill
    \begin{minipage}[t]{0.48\textwidth}
        \rule{0pt}{0pt}
        \centering
        \centering
\setlength{\tabcolsep}{6pt}
\renewcommand{\arraystretch}{1.1}

\caption{Ablation of connector types and token numbers, reporting accuracy at different spatial scales (WSI/ROI/Patch).}

\resizebox{\textwidth}{!}{
\begin{tabular}{l c c c c}
\toprule
\textbf{Connector} & \textbf{Token Num} & \textbf{WSI} & \textbf{ROI} & \textbf{Patch} \\
\midrule

wsi-llava & all & 81.61 & 88.05 & 86.83 \\
mlp & all & 83.33 & 87.24 & 83.66 \\
mlp & 32 & 85.34 & \textbf{89.10} & 86.13 \\
mlp & 1 & 85.34 & 87.47 & 87.25 \\
qformer & 500 & \textbf{89.37} & 88.40 & 85.25 \\
qformer & 32 & \underline{88.51} & \underline{88.75} & \textbf{88.10} \\
qformer & 1 & 87.64 & 88.52 & \underline{86.65} \\

\bottomrule
\end{tabular}
}
\label{tab:ablation_tokens}

    \end{minipage}
\end{table}

To validate the effectiveness of our architectural design, we conducted ablation studies on the HepatoPathoBench with identical hyperparameters.

\noindent\textbf{Effectiveness of Connector and Training Strategy.}
We compared our Q-Former-based connector against the MLP used in baseline methods like WSI-LLaVA~\cite{liang2025wsi}. As shown in Table~\ref{tab:ablation_connector}, the Q-Former architecture significantly outperforms the MLP baseline across all scales (+5.18\% on WSI). It also exhibits superior stability ($std=0.26$) compared to the MLP ($std=1.77$), validating its effectiveness in resampling variable-length visual sequences into fixed query embeddings. Furthermore, the results indicate that the Two-Stage setting yields performance gains (+2.76\% on ROI) compared to direct fine-tuning. We also report the performance of the MLP connector initialized from WSI-LLaVA weights, which benefits from broader pre-training and serves as a strong baseline.

\noindent\textbf{Analysis of Feature Redundancy and Token Efficiency.}
We hypothesize (RQ2) that standard slide encoders generate excessively redundant representations from gigapixel WSIs. To validate this, we evaluated the impact of token quantity on model performance. As detailed in Table~\ref{tab:ablation_tokens}, increasing the token count introduces noise and degrades performance. The Q-Former with 32 learnable queries achieves peak accuracy (88.75\% on ROI, 88.10\% on Patch), outperforming the 500-token configuration. A similar trend is observed in the MLP baseline, where even single-token(via max pooling) surpasses the use of all pack tokens. These results corroborate the sparsity of diagnostic signals in WSIs and underscore the necessity of condensing visual features into pack summaries. While the 500-query Q-Former excels in handling the extended sequences typical of WSIs, we select the 32-query setting to strike an optimal balance between inference efficiency and robust performance across varying spatial scales.

\section{Conclusion}
We present Hepato-LLaVA, a specialized MLLM for hepatocellular pathology analysis. To mitigate information redundancy in WSIs, we propose a Sparse Topo-Pack Attention mechanism that explicitly models 2D tissue topology. We also establish HepatoPathoVQA, a comprehensive multi-scale VQA dataset specifically for HCC WSIs. Extensive experiments demonstrate that our model achieves state-of-the-art performance, validating that topology-aware representations outperform high-dimensional redundant features. We believe this work demonstrates the efficacy of embedding pathological priors into deep learning frameworks, advancing the frontier of efficient AI for precision pathology.

%
%
%
\bibliographystyle{splncs04}
\bibliography{ref}

@article{weinstein2013cancer,
  title={The cancer genome atlas pan-cancer analysis project},
  author={Weinstein, John N and Collisson, Eric A and Mills, Gordon B and Shaw, Kenna R and Ozenberger, Brad A and Ellrott, Kyle and Shmulevich, Ilya and Sander, Chris and Stuart, Joshua M},
  journal={Nature genetics},
  volume={45},
  number={10},
  pages={1113--1120},
  year={2013},
  publisher={Nature Publishing Group}
}

@inproceedings{he2020momentum,
  title={Momentum contrast for unsupervised visual representation learning},
  author={He, Kaiming and Fan, Haoqi and Wu, Yuxin and Xie, Saining and Girshick, Ross},
  booktitle={Proceedings of the IEEE/CVF conference on computer vision and pattern recognition},
  pages={9729--9738},
  year={2020}
}

@inproceedings{li2023blip,
  title={Blip-2: Bootstrapping language-image pre-training with frozen image encoders and large language models},
  author={Li, Junnan and Li, Dongxu and Savarese, Silvio and Hoi, Steven},
  booktitle={International conference on machine learning},
  pages={19730--19742},
  year={2023},
  organization={PMLR}
}

@inproceedings{zhang2023huatuogpt,
  title={Huatuogpt, towards taming language model to be a doctor},
  author={Zhang, Hongbo and Chen, Junying and Jiang, Feng and Yu, Fei and Chen, Zhihong and Chen, Guiming and Li, Jianquan and Wu, Xiangbo and Zhiyi, Zhang and Xiao, Qingying and others},
  booktitle={Findings of the association for computational linguistics: EMNLP 2023},
  pages={10859--10885},
  year={2023}
}

@inproceedings{seyfioglu2024quilt,
  title={Quilt-llava: Visual instruction tuning by extracting localized narratives from open-source histopathology videos},
  author={Seyfioglu, Mehmet Saygin and Ikezogwo, Wisdom O and Ghezloo, Fatemeh and Krishna, Ranjay and Shapiro, Linda},
  booktitle={Proceedings of the IEEE/CVF Conference on Computer Vision and Pattern Recognition},
  pages={13183--13192},
  year={2024}
}

@inproceedings{song2024morphological,
  title={Morphological prototyping for unsupervised slide representation learning in computational pathology},
  author={Song, Andrew H and Chen, Richard J and Ding, Tong and Williamson, Drew FK and Jaume, Guillaume and Mahmood, Faisal},
  booktitle={Proceedings of the IEEE/CVF Conference on Computer Vision and Pattern Recognition},
  pages={11566--11578},
  year={2024}
}

@article{fourkioti2023camil,
  title={CAMIL: Context-aware multiple instance learning for cancer detection and subtyping in whole slide images},
  author={Fourkioti, Olga and De Vries, Matt and Jin, Chen and Alexander, Daniel C and Bakal, Chris},
  journal={arXiv preprint arXiv:2305.05314},
  year={2023}
}

@inproceedings{chen2025slidechat,
  title={Slidechat: A large vision-language assistant for whole-slide pathology image understanding},
  author={Chen, Ying and Wang, Guoan and Ji, Yuanfeng and Li, Yanjun and Ye, Jin and Li, Tianbin and Hu, Ming and Yu, Rongshan and Qiao, Yu and He, Junjun},
  booktitle={Proceedings of the Computer Vision and Pattern Recognition Conference},
  pages={5134--5143},
  year={2025}
}

@inproceedings{wu2025learning,
  title={Learning Heterogeneous Tissues with Mixture of Experts for Gigapixel Whole Slide Images},
  author={Wu, Junxian and Chen, Minheng and Ke, Xinyi and Xun, Tianwang and Jiang, Xiaoming and Zhou, Hongyu and Shao, Lizhi and Kong, Youyong},
  booktitle={Proceedings of the Computer Vision and Pattern Recognition Conference},
  pages={5144--5153},
  year={2025}
}

@article{zhang2025patho,
  title={Patho-R1: A Multimodal Reinforcement Learning-Based Pathology Expert Reasoner},
  author={Zhang, Wenchuan and Zhang, Penghao and Guo, Jingru and Cheng, Tao and Chen, Jie and Zhang, Shuwan and Zhang, Zhang and Yi, Yuhao and Bu, Hong},
  journal={arXiv preprint arXiv:2505.11404},
  year={2025}
}

@article{xu2025lingshu,
  title={Lingshu: A Generalist Foundation Model for Unified Multimodal Medical Understanding and Reasoning},
  author={Xu, Weiwen and Chan, Hou Pong and Li, Long and Aljunied, Mahani and Yuan, Ruifeng and Wang, Jianyu and Xiao, Chenghao and Chen, Guizhen and Liu, Chaoqun and Li, Zhaodonghui and others},
  journal={arXiv preprint arXiv:2506.07044},
  year={2025}
}

@article{zhuang2025libra,
  title={Libra-MIL: Multimodal Prototypes Stereoscopic Infused with Task-specific Language Priors for Few-shot Whole Slide Image Classification},
  author={Zhuang, Zhenfeng and Zhou, Fangyu and Wang, Liansheng},
  journal={arXiv preprint arXiv:2511.07941},
  year={2025}
}

@article{yan2025medreasoner,
  title={Medreasoner: Reinforcement learning drives reasoning grounding from clinical thought to pixel-level precision},
  author={Yan, Zhonghao and Diao, Muxi and Yang, Yuxuan and Jing, Ruoyan and Xu, Jiayuan and Zhang, Kaizhou and Yang, Lele and Liu, Yanxi and Liang, Kongming and Ma, Zhanyu},
  journal={arXiv preprint arXiv:2508.08177},
  year={2025}
}

@inproceedings{liang2025wsi,
  title={Wsi-llava: A multimodal large language model for whole slide image},
  author={Liang, Yuci and Lyu, Xinheng and Chen, Wenting and Ding, Meidan and Zhang, Jipeng and He, Xiangjian and Wu, Song and Xing, Xiaohan and Yang, Sen and Wang, Xiyue and others},
  booktitle={Proceedings of the IEEE/CVF International Conference on Computer Vision},
  pages={22718--22727},
  year={2025}
}

@misc{google_gemini_2026,
  author = {{Google}},
  title = {Gemini 3 Flash (Free Tier)},
  year = {2026},
  howpublished = {\url{https://gemini.google.com/}},
  note = {Accessed: 2026-02-02}
}

@misc{hcmi,
  title        = {Human Cancer Models Initiative (HCMI)},
  author       = {{National Cancer Institute} and {National Human Genome Research Institute} and {European Bioinformatics Institute}},
  year         = {2017},
  howpublished = {\url{www.cancer.gov/ccg/research/functional-genomics/hcmi}},
  note         = {Accessed: 2026-02-03}
}

@inproceedings{banerjee2005meteor,
  title={METEOR: An automatic metric for MT evaluation with improved correlation with human judgments},
  author={Banerjee, Satanjeev and Lavie, Alon},
  booktitle={Proceedings of the acl workshop on intrinsic and extrinsic evaluation measures for machine translation and/or summarization},
  pages={65--72},
  year={2005}
}

\end{document}